\documentclass{article}
\PassOptionsToPackage{numbers, compress}{natbib}

\usepackage[final]{neurips_2024}

\usepackage[utf8]{inputenc}
\usepackage[T1]{fontenc}
\usepackage{hyperref}
\usepackage{url}
\usepackage{booktabs}
\usepackage{amsfonts}
\usepackage{nicefrac}
\usepackage{microtype}
\usepackage{xcolor}
\usepackage{graphicx}
\usepackage{graphbox}
\usepackage{multirow}
\usepackage{makecell}
\usepackage{wrapfig}
\usepackage{amsmath}
\usepackage{enumitem}
\usepackage{adjustbox}

\title{Self-supervised Transformation Learning \\ for Equivariant Representations}

\author{%
  Jaemyung Yu$^{1}$\quad Jaehyun Choi$^{1}$\quad Dong-Jae Lee$^{1}$\quad HyeongGwon Hong$^{1}$\quad Junmo Kim$^{1}$\vspace{0.5mm}\\
  $^{1}$Korea Advanced Institute of Science and Technology (KAIST)\vspace{0.5mm}\\
  \texttt{\{jaemyung,chlwogus,jhtwosun,honggudrnjs,junmo.kim\}@kaist.ac.kr}\\
}

\begin{document}

\maketitle

\begin{abstract}
Unsupervised representation learning has significantly advanced various machine learning tasks. In the computer vision domain, state-of-the-art approaches utilize transformations like random crop and color jitter to achieve invariant representations, embedding semantically the same inputs despite transformations. However, this can degrade performance in tasks requiring precise features, such as localization or flower classification. To address this, recent research incorporates equivariant representation learning, which captures transformation-sensitive information. However, current methods depend on transformation labels and thus struggle with interdependency and complex transformations. We propose Self-supervised Transformation Learning (STL), replacing transformation labels with transformation representations derived from image pairs. The proposed method ensures transformation representation is image-invariant and learns corresponding equivariant transformations, enhancing performance without increased batch complexity. We demonstrate the approach’s effectiveness across diverse classification and detection tasks, outperforming existing methods in 7 out of 11 benchmarks and excelling in detection. By integrating complex transformations like AugMix, unusable by prior equivariant methods, this approach enhances performance across tasks, underscoring its adaptability and resilience. Additionally, its compatibility with various base models highlights its flexibility and broad applicability. The code is available at \url{https://github.com/jaemyung-u/stl}.
\end{abstract}

\subsection*{Keyword}
Equivariant Learning, Transformation Representation, Self-supervised Transformation Learning

\section{Introduction}
\label{introduction}
Recently, unsupervised representation learning~\citep{rotnet, ssl_old_1, ssl_old_2} has made remarkable strides in various machine learning tasks and is actively employed as the ground model. In particular to the state-of-the-art computer vision models~\citep{simclr, byol, swav, simsiam, barlow, vicreg}, invariant representation learning utilizes various augmentations, hereinafter \textit{transformations}, including but not limited to random crop, horizontal flip, color jitter, and Gaussian blur. Their objective is to embed the semantically same inputs obtained through transformation, based on the notion that semantic differences due to the various transformations are inconsequential. Unfortunately, although effective most of the time, it does not always guarantee performance gain in the target downstream tasks. For instance, in tasks such as localization or flower classification, applying random crop or color jitter for invariant learning degrades performance by diluting discriminative features regarding the position of the object or the color of the flower, respectively.

\begin{figure}[t]
    \centering
    \includegraphics[width=1.0\linewidth]{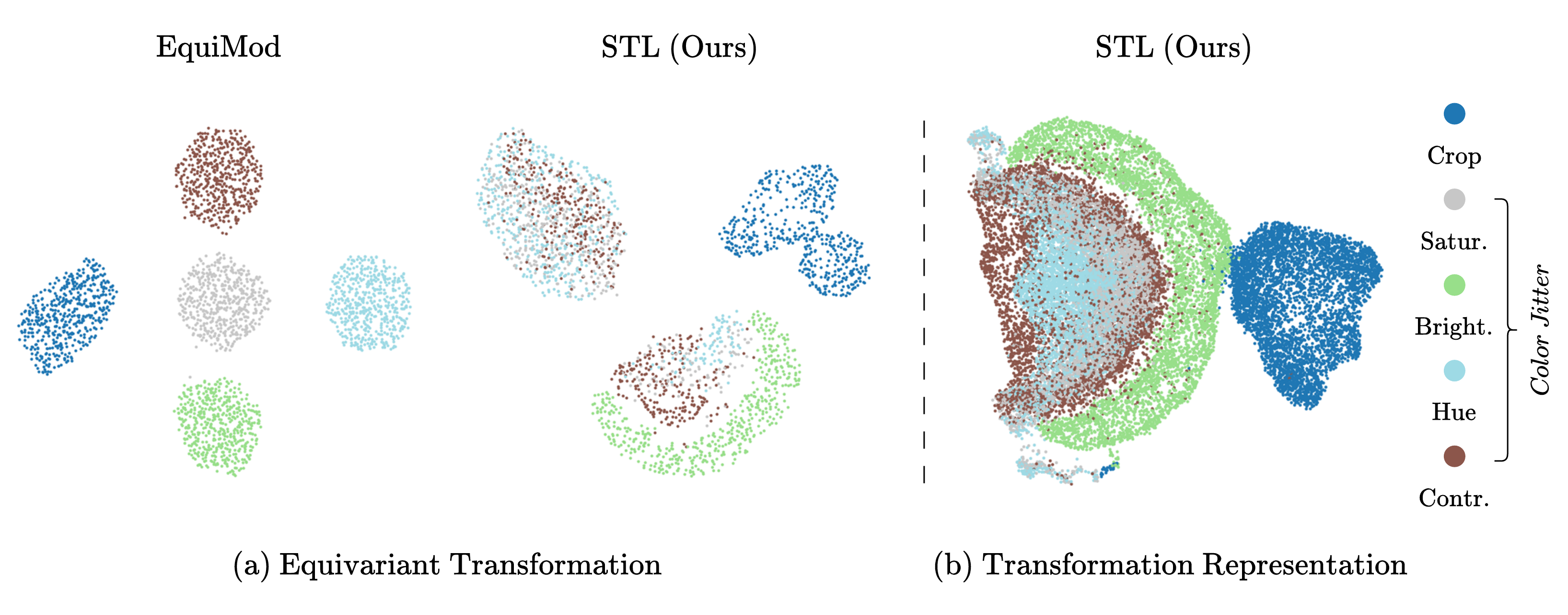}
    \caption{\small{\textbf{Visualization of Equivariant Transformation and Transformation Representation.} (Left) UMAP \cite{umap} visualizations of functional weights from equivariant transformations implemented with a hypernetwork. EquiMod uses transformation labels to generate these weights, while STL derives them from the representation pairs of transformed and original image. (Right) UMAP visualizations of transformation representations obtained from representation pairs of original input image and transformed input image.}}
    \vspace{-1.5em}
    \label{fig:transformation_representation}
\end{figure}

To circumvent such a problem while harnessing the benefits of invariant representation learning, the research has transitioned towards incorporating equivariant representation learning~\citep{e-ssl, augself, sen, equimod, sie} alongside invariant representation learning. As learning an equivariant representation in parallel with an invariant representation requires the representation to be responsive to transformations, the objective of equivariant representation learning is to capture transformation-sensitive information. This can be achieved either by training the model to predict transformations through representation pairs as in E-SSL~\citep{e-ssl} and AugSelf~\citep{augself}, referred to as \textit{implicit equivariant learning}, or by learning the corresponding equivariant transformations in the representation space as in SEN~\citep{sen}, EquiMod~\citep{equimod}, and SIE~\citep{sie}, referred to as \textit{explicit equivariant learning}. However, these methods face a major issue due to the requirement of transformation labels. Firstly, optimizing with a transformation label, each transformation is treated independently, disregarding interdependency among transformations. Consequently, each component in color jitter transformation is treated distinctively although they are related to each other in the sense that they are applying transformation in the color as shown in Figure \ref{fig:transformation_representation} (a). Secondly, due to the limited expressiveness of transformation labels, it fails to encompass complex transformations such as AugMix~\citep{augmix}, where the sequences of transformations are randomly applied with varying weights. After all, the reliance on the transformation label limits the performance gain in equivariant representation learning.

To address the aforementioned limitations, we propose Self-supervised Transformation Learning (STL) for learning representation of transformation, which enhances the potential of equivariant learning. In STL, the transformation label is replaced by the transformation representation derived from the representation pairs of original and transformed image. To ensure the transformation representation captures the information of transformation, we train it to be invariant to the same transformation applied to various images, similar to how contrastive learning trains image representation to be invariant to various transformations. Based on the derived transformation representations, we aim to learn the corresponding equivariant transformations in the representation space. To avoid the \textit{trivial solution} when learning the equivariant transformation, we apply the transformation representation obtained from another image but with the same transformation. Through optimization, STL is able to learn the transformation representation with the same batch complexity as previous methods.

We demonstrate the effectiveness of STL in transfer learning across various datasets and tasks, including both classification and detection. Additionally, by incorporating AugMix, a complex transformation that is not feasible with existing equivariant learning, STL enhances performance across all tasks, showcasing its broad applicability. STL’s compatibility with diverse base models, further highlights its versatility, as it achieves the highest average accuracy across foundational models.  Extensive experiments and ablation studies further validate STL’s ability to capture interdependencies among transformations in an unsupervised manner. This is evident not only in transformation representations and equivariant transformation clustering, which reveal nuanced relationships between transformations (see Figure \ref{fig:transformation_representation}), but also in its superior performance in transformation prediction, surpassing existing equivariant learning methods.

\section{Preliminaries: Transformation Invariant and Equivariant Learning}
\label{preliminary}

\textbf{Transformation as Group Action.} A group \(G\) consists of elements and an operation that satisfies closure, associativity, the existence of an identity element \(e\), and the existence of inverses for all elements. The group action of \(G\) on a set \(X\) is defined as a function \(\cdot: G \times X \rightarrow X\), which ensures the identity operation \(e \cdot x = x\) \ for all \(x \in X\) and maintaining the compatibility of the operations \((g \cdot h) \cdot x = g \cdot (h \cdot x)\) \ for all \(g, h \in G\). In the context of image processing, transformations can be viewed as a group action where the group \(T\) consists of transformations, and the set \(X\) is the collection of images. For example, if \(R \in T\) is a rotation and \(F \in T\) is a flip transformation, applying \(R\) followed by \(F\) to an image \(x \in X\) is represented as \(F \cdot (R \cdot x) = F(R(x))\). This structure ensures that image transformations are consistent and reversible.

\textbf{Transformation Invariant Representation.} In the context of transformation as group action, the \textit{transformation invariance} of an image representation obtained through an encoder \(f:X \rightarrow Y\), which maps the set of images \(X\) to representation space \(Y\), is defined as follows: Let \(T\) be a group of transformations applied to an input image \(x \in X\), with \(t(x)\) representing the image after transformation \(t \in T\). The representation \(f(x)\) is invariant to all transformations in \(T\) if \(f(x)\) of an input image \(x\) remains unchanged even after applying any transformation \(t\) to \(x\).
\begin{equation}
    \label{eq:transformation-invariant-representation}
    f(x) = f(t(x)) \quad \forall t \in T.
\end{equation}
In self-supervised learning, leveraging transformation invariance is crucial for learning representations, as it aligns transformed input images within the representation space. The objective of transformation invariant representation learning \(\mathcal{L}_\text{inv}\) is formalized with a dissimilarity loss \(\mathcal{L}\) as follows:
\begin{equation}
    \label{eq:transformation-invariant-learning}
    \min_{f} \, \mathbb{E}_{x, t}\Big[\mathcal{L}_{\text{inv}}(x, t)\Big] \quad \text{s.t.} \quad \mathcal{L}_{\text{inv}}(x, t) = \mathcal{L}(f(x), f(t(x))),
\end{equation}
where \(\mathcal{L}\) represents the dissimilarity metric between representations. In contrastive learning, \(\mathcal{L}\) is instantiated as metrics like the InfoNCE \citep{infonce} loss, which reduces the distance between representations of semantically similar inputs while increasing the gap between those of dissimilar inputs.

\textbf{Transformation Equivariant Representation.} Extending the concept of invariance, \textit{transformation equivariance} ensures that an image representation changes predictably according to the applied transformation. The representation \(f(x)\) is equivariant to all transformations in \(T\) if there exists a group action \(\phi: T \times Y \rightarrow Y\) on the representation space \(Y\), ensuring that the application of \(t\) to an image \(x\) leads to a corresponding transformation \(\phi(t, f(x))\) in its representation.
\begin{equation}
    \label{eq:transformation-equivariant-representation}
    f(t(x)) = \phi(t, f(x)) \quad \forall t \in T.
\end{equation}
The function \(\phi(t, \cdot): Y \rightarrow Y\) is referred to as the equivariant transformation on the representation space corresponding to the transformation \(t\). When the equivariant transformation \(\phi(t, \cdot)\) becomes the identity, it implies transformation invariance, showcasing the model's insensitivity to transformations. In EquiMod \citep{equimod} and SIE \citep{sie}, equivariant transformation \(\phi(\hat{t}, \cdot)\) is represented by a network parameterized by the corresponding transformation label \(\hat{t}\). This network is trained to associate each transformation \(t\) with its corresponding equivariant transformation \(\phi(\hat{t}, \cdot)\) in the representation space. The objective of transformation equivariant representation learning \(\mathcal{L}_\text{equi}\) is formalized as follows:
\begin{equation}
    \label{eq:transformation-equivariant-learning}
    \min_{f, \phi} \, \mathbb{E}_{x, t}\Big[\mathcal{L}_{\text{equi}}(x, t)\Big] \quad \text{s.t.} \quad \mathcal{L}_{\text{equi}}(x, t) = \mathcal{L}(\phi(\hat{t}, f(x)), \ f(t(x))).
\end{equation}

\section{Equivariant Learning with Self-supervised Transformation Learning}
\label{method}

\begin{figure}[tb]
    \centering
    \includegraphics[width=\textwidth]{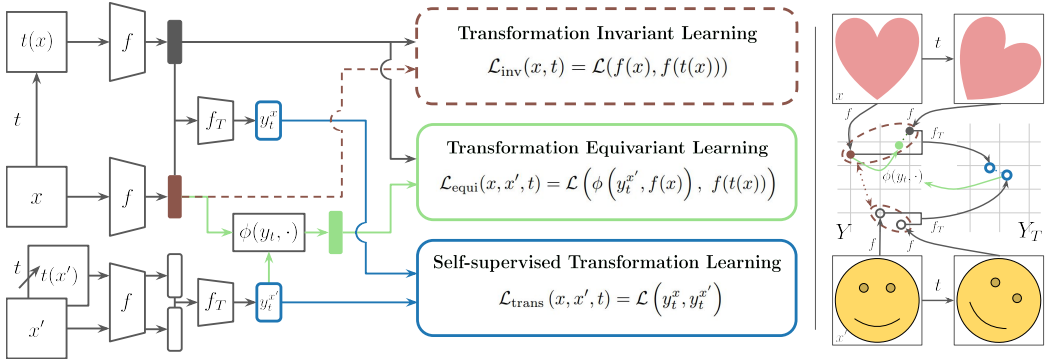}
    \caption{\small{\textbf{Transformation Equivariant Learning with Self-supervised Transformation Learning.} (Left) The overall framework of STL. For given image and transformations, it demonstrates: 1) transformation invariant learning, which aligns the representations of image and transformed image; 2) transformation equivariant learning, where the representation of image transformed by an equivariant transformation (obtained from the transformation representation of different image with the same applied transformation) aligns with the transformed image's representation; 3) self-supervised transformation learning, which aligns the transformation representations obtained from different image pairs. (Right) It illustrates the transformations of each representation and the equivariant transformations within the representation space.}}
    \label{fig:overview}
\end{figure}

\subsection{Equivariant Learning without Transformation Label}
\textbf{Transformation Representation.} Instead of relying on transformation labels that require knowledge of the transformation group structure, we propose leveraging pairs of representations derived from original images and their transformed counterparts to implicitly represent transformations. We introduce an auxiliary encoder \(f_T: Y \times Y \rightarrow Y_T\) designed to process pairs of representations \( (f(x), f(t(x)))\), where \( f(x) \) and \( f(t(x)) \) are the representation of the original image and of the transformed image respectively. This encoder outputs a transformation representation \(y^x_t \in Y_T\), capturing the inherent transformation between the original and transformed images without explicit transformation labels. 
\begin{equation}
    \label{eq:transfomation-representation}
    y^x_t = f_T(f(x), \ f(t(x))) \in Y_T \quad \text{for} \ t \in T \ \text{and} \ x \in X.
\end{equation}

\textbf{Equivariant Learning with Transformation Representation.} In contrast to methods directly utilizing transformation labels to learn equivariant transformations in the representation space, our proposed approach substitutes labels with transformation representations derived from pairs of images. However, there is a risk of encountering a trivial solution when using the transformation representation derived from the representation pair of the same image for which the equivariant transformation is applied. Specifically, the equivariant transformation might simply output the same representation \( f(t(x)) = \phi(f_T(f(x), f(t(x))), f(x)) \) that was used to obtain the transformation representation. To address this issue, we propose using transformation representations \(y^{x'}_t\)derived from pairs of a different image \( x' \) to apply equivariant transformations. Therefore, the transformed representation of the image through equivariant transformation in the representation space can be expressed as follows:
\begin{equation}
    \label{eq:equivariant_transformed_representation}
    \phi\left(y^{x'}_t, \ f(x)\right) = \phi\left(f_T\left(f\left(x'\right), f\left(t(x')\right)\right), \ f(x)\right) \quad \text{for} \ x \neq x' \in X.
\end{equation}
Leveraging this approach, it is possible to learn transformation equivariant representations without explicit transformation labels, through the following objective:
\begin{equation}
    \label{eq:transformation-equivariant-learning-wo-labels}
    \min_{f, f_T, \phi} \, \mathbb{E}_{x \neq x', t} \Big[\mathcal{L}_{\text{equi}}(x, x', t)\Big] \quad \text{s.t.} \quad \mathcal{L}_{\text{equi}}(x, x', t) = \mathcal{L}\left(\phi\left(y^{x'}_t, f(x)\right), \ f(t(x))\right).
\end{equation}

\subsection{Self-supervised Transformation Learning (STL)}
We hypothesize that transformation representations, \(y^x_t\), derived from an input image \(x\) and its transformed image \(t(x)\), encode the transformation \(t\) independently of the input image \(x\). Nonetheless, ensuring image-invariant encoding of \(t\) is not trivial.
\begin{equation}
    \label{eq:sample-invariant-transformation-representation}
    y^x_t = y^{x'}_t \quad \forall x \neq x' \in X.
\end{equation}
To address this, we introduce Self-supervised Transformation Learning (STL), which adapts contrastive learning for transformation representation. Contrary to contrastive learning, which aims to align representations \(f(x)\) of the same image under different transformations to promote transformation invariance, STL instead focuses on aligning transformation representations, \(y^x_t\) and \(y^{x'}_t\), derived from different images \(x \neq x'\) subjected to the identical transformation \(t\). The objective of STL is formalized as follows:
\begin{equation}
    \label{eq:self-supervised-transformation-learning}
    \min_{f, f_T} \, \mathbb{E}_{x \neq x', t}\Big[\mathcal{L}_{\text{trans}}\left(x, x', t\right)\Big] \quad \text{s.t.} \quad \mathcal{L}_{\text{trans}}\left(x, x', t\right) = \mathcal{L}\left(y^x_t, y^{x'}_t\right).
\end{equation} 
Hereinafter, equivariant learning using transformation representations learned through self-supervised transformation learning will simply be referred to as STL.

\begin{figure}[tb]
    \centering
    \includegraphics[width=\textwidth]{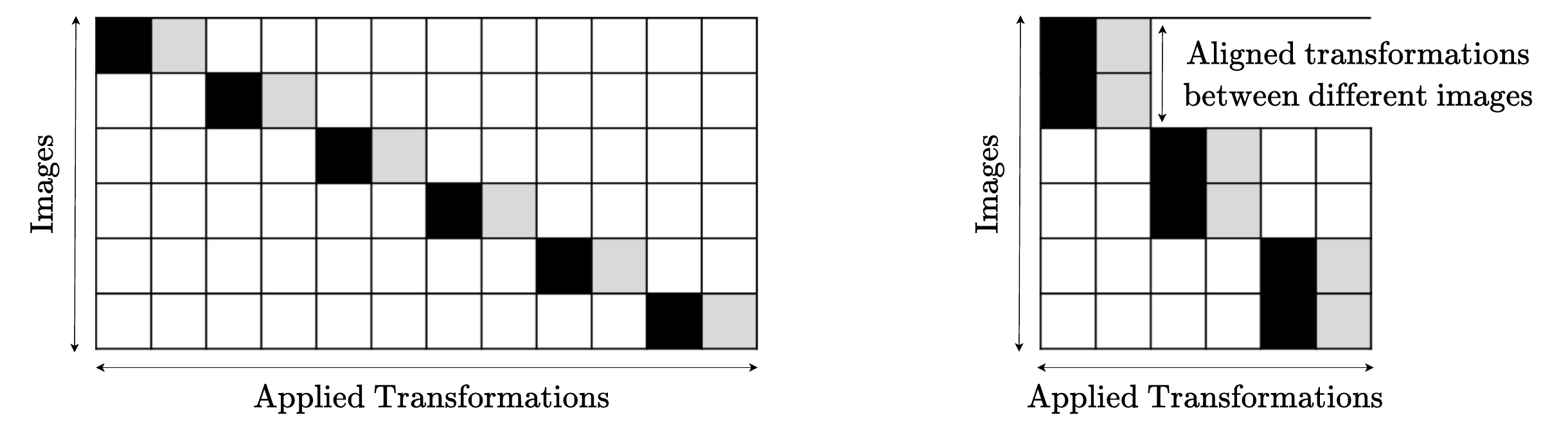}
    \caption{\small{\textbf{Aligned Transformed Batch.} (Left) In self-supervised learning methods, batch compositions typically involve applying two different transformations to each input image. (Right) In STL, batches are composed by pairing two images together, and applying the same transformation pair.}}
    \vspace{-1em}
    \label{fig:aligned_transformed_batch}
\end{figure}

\subsection{Implementation Details}
\textbf{Dissimilarity Metric.}  We use the InfoNCE loss from SimCLR~\citep{simclr} for the formulation and implementation of STL. Our methodology is not limited to this model, as demonstrated by ablation study, which shows the feasibility of applying our approach across various self-supervised learning models, such as BYOL~\citep{byol}, SimSiam~\citep{simsiam}, and Barlow Twins~\citep{barlow}, through straightforward extensions (see Appendix \ref{section:supp:various_base_models}). Like EquiMod~\citep{equimod}, we employ specialized projectors \(g_\text{inv}\), \(g_\text{equi}\), and \(g_\text{trans}\) to map representations into distinct embedding spaces \(Z_\text{inv}\), \(Z_\text{equi}\), and \(Z_\text{trans}\), aligned with the objectives of invariant, equivariant, and transformation representation learning. We adopt the InfoNCE loss as the dissimilarity metric across these spaces, similar to the approach in SimCLR. The InfoNCE loss function is defined as follows:
\begin{equation}
    \label{eq:infonce-loss}
    \mathcal{L}_{\text{InfoNCE}}\left(y, y^+ ; g, \tau, \{y\}_i\right) = -\log \frac{\exp\left(\text{sim}\left(g(y), g(y^+)\right)/\tau\right)}{\sum_{y_i \neq y} \exp\left(\text{sim}\left(g(y), g(y_i)\right)/\tau\right)},
\end{equation}
where \(y\) represents representation of an input image, \(y^+\) denotes the corresponding representation to align, \(\text{sim}(\cdot)\) indicates a similarity function, and \(\tau\) is a temperature scaling parameter. For simplicity, batch \(\{y\}_i\) are omitted in the subsequent loss functions.
We define three specific loss functions for invariant, equivariant, and transformation representation learning, each building on the InfoNCE loss:
\begin{align}
    \label{eq:inv-loss}
    &\mathcal{L}_{\text{inv}}(x, t) \qquad =  \ \mathcal{L}_{\text{InfoNCE}}\big(f(x), f(t(x)); \ g_\text{inv}, \tau_\text{inv}\big), \\
    &\mathcal{L}_{\text{equi }}(x, x', t) \ = \ \mathcal{L}_{\text{InfoNCE}}\big(\phi\big(y^{x'}_t, f(x)\big), \ f(t(x)); \ g_\text{equi}, \tau_\text{equi}\big), \\
    &\mathcal{L}_{\text{trans}}(x, x', t) \ = \ \mathcal{L}_{\text{InfoNCE}}\big(y^x_t, \ y^{x'}_t; \ g_\text{trans}, \tau_\text{trans}\big).
\end{align}

\setlength{\intextsep}{0pt}
\setlength{\abovecaptionskip}{0pt}
\setlength{\belowcaptionskip}{8pt}

\begin{wraptable}{R}{0.3\textwidth}
    \scriptsize
    \centering
    \caption{\small{\textbf{Computational Cost.}} Forward-backward time per iteration on NVIDIA 3090 GPU with ResNet-50 and batch size 256.}
    \label{tab:computational_cost}
    \resizebox{\linewidth}{!}{
        \begin{tabular}{lcc}
        \toprule
        \textbf{Method} & \textbf{Time (s)} & \textbf{Ratio} \\
        \midrule
        SimCLR & 0.51 & 1.00 \\
        AugSelf & 0.51 & 1.00 \\
        EquiMod & 0.53 & 1.01 \\
        \textbf{STL (Ours)} & 0.56 & 1.11 \\
        \bottomrule
        \end{tabular}
    }
\end{wraptable}

\textbf{Aligned Transformed Batch.} To implement STL, we need transformation representations obtained from different images. Unlike typical batch configurations in contrastive learning, where different transformations are applied to each input image, we construct batches by applying identical transformations to image pairs, as illustrated in Figure~\ref{fig:aligned_transformed_batch}. In our approach, each image undergoes two distinct transformations, denoted \(t\) and \(t'\), but for simplicity, we consider only a single direction of transformation, treating \(t\) as equivalent to \(t' \cdot t^{-1}\). This aligned transformed batch configuration maintains the same computational complexity as typical contrastive learning setups while preserving input diversity. It also increases the count of identical transformations applied across different images, which is essential for transformation learning, without diminishing input diversity. To assess computational costs, we measured forward-backward time over 1000 iterations following a 1000-iteration warm-up. With an auxiliary network and loss calculation, our approach required only about 10\% more time per iteration than SimCLR, which focuses solely on invariant learning, as shown in Table~\ref{tab:computational_cost}.

\textbf{Overall Objective.} The overall framework of STL is shown in Figure~\ref{fig:overview}. Using aligned transformed batch, along with the InfoNCE loss, we define the overall objective with the hyperparameters \(\lambda_\text{inv}\), \(\lambda_\text{equi}\) and \(\lambda_\text{trans}\) for balancing the respective losses as follows:
\begin{equation}
    \label{eq:overall-objective}
    \min_{f, f_T, \phi} \, \mathbb{E}_{x \neq x', t} \Big[ \lambda_\text{inv} \mathcal{L}_\text{inv}(x, t) + \lambda_\text{equi} \mathcal{L}_\text{equi}(x, x', t) + \lambda_\text{trans} \mathcal{L}_\text{trans}(x, x', t) \Big].
\end{equation}

\section{Experiments}
\label{experiments}

\begin{table}[t]
    \renewcommand{\arraystretch}{1.1}
    \centering
    \caption{\small{\textbf{Out-domain Classification.} Evaluation of representation generalizability on the out-domain downstream classification tasks. Linear evaluation accuracy (\%) is reported for ResNet-50 pretrained on ImageNet100.}}
    \label{tab:out_domain_classification}
    \resizebox{\linewidth}{!}{
        \begin{tabular}{lccccccccccc|c}
        \toprule
        \textbf{Method} & \textbf{CIFAR10} & \textbf{CIFAR100} & \textbf{Food} & \textbf{MIT67} & \textbf{Pets} & \textbf{Flowers} & \textbf{Caltech101} & \textbf{Cars} & \textbf{Aircraft} & \textbf{DTD} & \textbf{SUN397} & \textbf{Mean} \\
        
        \midrule
        \multicolumn{12}{l|}{\textit{\textcolor{gray}Invariant Learning} :} & \\
        SimCLR & 84.24 & 64.15 & 59.00 & 54.78 & 58.95 & 91.58 & 79.32 & 27.07 & 36.00 & 66.01 & 42.77 & 60.35 \\
        with AugMix & 86.90 & \textbf{67.70} & 62.90 & 57.24 & 63.75 & 93.16 & 83.67 & 32.37 & 43.17 & 67.93 & 46.15 & 64.09 \\
        
        \midrule
        \multicolumn{12}{l|}{\textit{\textcolor{gray}Implicit Equivariant Learning} :} \\
        E-SSL & 85.09 & 65.74 & 60.91 & 56.64 & 61.00 & 92.31 & 80.77 & 28.84 & 38.04 & 66.38 & 43.49 & 61.75 \\
        AugSelf & 85.55 & 66.09 & 62.63 & 57.16 & 62.61 & 93.41 & 82.33 & 30.71 & 40.35 & 68.51 & 45.24 & 63.14 \\
        
        \midrule
        \multicolumn{12}{l|}{\textit{\textcolor{gray}Explicit Equivariant Learning} :} \\
        SEN     & 80.68 & 56.53 & 52.50 & 46.79 & 45.27 & 79.24 & 73.42 & 14.41 & 27.51 & 57.45 & 33.51 & 51.57 \\
        EquiMod & 82.89 & 61.36 & 56.38 & 52.84 & 52.68 & 87.42 & 79.17 & 22.02 & 34.62 & 64.10 & 39.86 & 57.58 \\
        SIE & 81.72 & 58.49 & 54.04 & 49.70 & 47.21 & 84.37 & 74.39 & 16.71 & 31.68 & 59.20 & 35.29 & 53.89 \\
        
        \textbf{STL (Ours)} & 86.55 & 66.84 & 64.32 & 56.64 & 65.00 & 94.51 & 81.83 & 35.44 & 45.42 & 64.68 & 44.69 & 64.18 \\
        \textbf{with AugMix (Ours)} & \textbf{87.19} & \textbf{67.70} & \textbf{66.12} & \textbf{59.70} & \textbf{67.10} & \textbf{94.87} & \textbf{84.61} & \textbf{38.48} & \textbf{46.14} & \textbf{69.57} & \textbf{45.75} & \textbf{66.11} \\
        \bottomrule
        \end{tabular}
    }
\end{table}
\setlength{\intextsep}{0pt}
\setlength{\abovecaptionskip}{0pt}
\setlength{\belowcaptionskip}{8pt}

\begin{wraptable}{R}{0.35\textwidth}
    \scriptsize
    \centering
    \caption{\small{\textbf{In-domain Classification.} Evaluation of representation on in-domain classification task. Linear evaluation accuracy (\%) is reported for ResNet-50 pretrained on ImageNet100.}}
    \label{tab:in_domain_classification}
    \resizebox{\linewidth}{!}{
        \begin{tabular}{lcc}
        \toprule
        \textbf{Method} & \textbf{Accuracy} \\
        
        \midrule
        \multicolumn{2}{l}{\textit{\textcolor{gray}Invariant Learning} :} \\
        SimCLR & 81.20 \\
        SimCLR with AugMix & 80.54 \\
        
        \midrule
        \multicolumn{2}{l}{\textit{\textcolor{gray}Implicit Equivariant Learning} :} \\
        E-SSL & \textbf{82.10} \\
        AugSelf & 81.08 \\
        
        \midrule
        \multicolumn{2}{l}{\textit{\textcolor{gray}Explicit Equivariant Learning} :} \\
        SEN & 76.32 \\
        EquiMod & 80.70 \\
        SIE & 79.40 \\
        
        \textbf{STL (Ours)} & 81.10 \\
        \textbf{STL with AugMix (Ours)} & 81.64 \\
        \bottomrule
        \end{tabular}
    }
\end{wraptable}

\textbf{Baselines.} We compare STL with implicit and explicit equivariant learning methods, using SimCLR~\citep{simclr} as the base invariant model. Implicit methods (E-SSL~\citep{e-ssl} and AugSelf~\citep{augself}) learn equivariant representations via transformation prediction tasks. Explicit methods (SEN~\citep{sen}, EquiMod~\citep{equimod}, and SIE~\citep{sie}) use transformation labels for equivariant learning. All methods, including STL, are trained and evaluated with SimCLR as the base model. Experiments with other base models are included in the ablation study. 

\textbf{Datasets.} We pretrain on STL10~\citep{stl10} with ResNet-18 and ImageNet100~\citep{imagenet, imagenet_samplesplit} with ResNet-50, following the split in~\citep{imagenet_samplesplit}. Evaluation spans 11 downstream classification tasks (CIFAR10/100~\citep{cifar}, Food~\citep{food}, MIT67~\citep{mit67}, Pets~\citep{pets}, Flowers~\citep{flowers}, Caltech101~\citep{caltech101}, Cars~\citep{cars}, Aircraft~\citep{aircraft}, DTD~\citep{dtd}, SUN397~\citep{sun397}) with a linear transfer protocol~\citep{lineval}. For detection, VOC07+12 dataset~\citep{voc07} and the protocol from~\citep{barlow} are used. Dataset and protocol details are in Appendix~\ref{section:supp:dataset_info} and ~\ref{section:supp:evaluation_protocol}.

\textbf{Setup.} For STL and explicit baselines (SEN, EquiMod, and SIE), we use an equivariant transformation network with a hypernetwork based on SIE. In SEN, EquiMod, and SIE, the hypernetwork uses transformation labels; in STL, it leverages transformation representations. STL also includes a 3-layer MLP with a 512-dimensional hidden layer to encode 128-dimensional transformation representations from input pairs. Equivariant transformations include random crop and color jitter, with other transformations applied randomly, consistent with typical contrastive learning. The transformation prediction loss weight is set to 0.5 for implicit baselines, and the equivariant learning weight is set to 1 for explicit baselines. STL uses weights of 1, 1, and 0.2 for invariant, equivariant, and transformation learning losses, respectively. We apply AugMix~\citep{augmix} to evaluate STL’s adaptability to complex transformations, incompatible with other methods. Details on transformation labels in standard equivariant learning are in Appendix~\ref{section:supp:transformation_labels}. All analyses and ablations, except main experiments, use STL10-pretrained models. Additional setup details are in Appendix~\ref{section:supp:pretraining_setup}.

\subsection{Main Results}
\textbf{Image Classification.} To assess generalizability, we apply the linear evaluation protocol on various downstream tasks. As shown in Table~\ref{tab:out_domain_classification}, STL outperforms existing methods on 7 out of 11 datasets. With AugMix, a complex transformation combination, STL achieves the highest performance across all datasets, underscoring its ability to generalize across diverse transformations, even those without explicit labels, to improve generalization. In Table~\ref{tab:in_domain_classification}, STL shows a minimal trade-off on in-domain tasks compared to SimCLR, with only a slight decrease from 81.20\% to 81.10\%, a smaller trade-off than other explicit equivariant models. Combined with AugMix, STL reaches 81.64\%, showing adaptability to complex transformations and further enhancing in-domain performance. Results on STL10-pretrained models are provided in the Appendix~\ref{section:supp:stl10_results}.

\begin{table}[t]
    \tiny
    \centering
    \begin{minipage}[t]{0.48\textwidth}
        \centering
        \caption{\small{\textbf{Object Detection.} Evaluation of representation generalizability on a downstream object detection task. Average precision is reported for ImageNet100-pretrained ResNet-50 fine-tuned on VOC07+12.}}
        \label{tab:object_detection}
        \resizebox{\linewidth}{!}{
            \begin{tabular}{lccc}
            \toprule
            \textbf{Method} & \textbf{AP$_{\mathrm{all}}$} & \textbf{AP$_{50}$} & \textbf{AP$_{75}$} \\
            
            \midrule
            SimCLR & 45.67 & 72.50 & 47.83 \\
            AugSelf & 45.99 & 72.46 & 49.23 \\
            EquiMod & 51.55 & 78.03 & 56.17 \\
            
            \midrule
            \textbf{STL (Ours)} & 51.95 & 78.34 & 56.96 \\
            \textbf{with AugMix (Ours)} & \textbf{52.70} & \textbf{78.81} & \textbf{57.76} \\
            \bottomrule
            \end{tabular}
        }
    \end{minipage}
    \hfill
    \begin{minipage}[t]{0.48\textwidth}
        \renewcommand{\arraystretch}{1.09}
        \centering
        \caption{\small{\textbf{Transformation Prediction.} Evaluation of transformation representation from learned represetation pairs. Regression tasks use MSE loss, and transformation type classification uses accuracy (\%).}}
        \label{tab:transformation_prediction}
        \resizebox{\linewidth}{!}{ 
            \begin{tabular}{lcccccc}
            \toprule
            & \multicolumn{3}{c}{\textbf{Regression ($\downarrow$)}} & \multicolumn{1}{c}{\textbf{Classification ($\uparrow$)}} \\
            \cmidrule(lr){2-4} \cmidrule(lr){5-5}
            \textbf{Method} & Crop & Color & All & Trans. Type \\
            \midrule
            
            SimCLR & 0.02 & 0.13 & 0.08 & 68.54 \\
            AugSelf & \textbf{0.01} & 0.04 & 0.03 & 88.49 \\
            EquiMod & \textbf{0.01} & 0.07 & 0.04 & 82.20 \\
            
            \midrule
            \textbf{STL (Ours) }& \textbf{0.01} &\textbf{ 0.03} & \textbf{0.02} & \textbf{93.67} \\
            \bottomrule
            \end{tabular}
        }
    \end{minipage}
\end{table}
\textbf{Object Detection.} We evaluate STL on the VOC07+12 object detection task. As shown in Table~\ref{tab:object_detection}, STL outperforms the invariant learning model SimCLR, as well as AugSelf, a representative model for implicit equivariant learning, and EquiMod, a representative for explicit equivariant learning, across all metrics: AP$_{\mathrm{all}}$, AP$_{\mathrm{50}}$ and AP$_{\mathrm{75}}$. STL achieves 51.95 in AP$_{\mathrm{all}}$, and with AugMix, further improves to 52.70 in AP$_{\mathrm{all}}$, 78.81 in AP$_{\mathrm{50}}$, and 57.76 in AP$_{\mathrm{75}}$, demonstrating robust adaptability to complex transformations and enhanced precision in localization.

\subsection{Analysis}
\begin{wrapfigure}{R}{0.355\textwidth}
    \centering
    \includegraphics[width=0.355\textwidth]{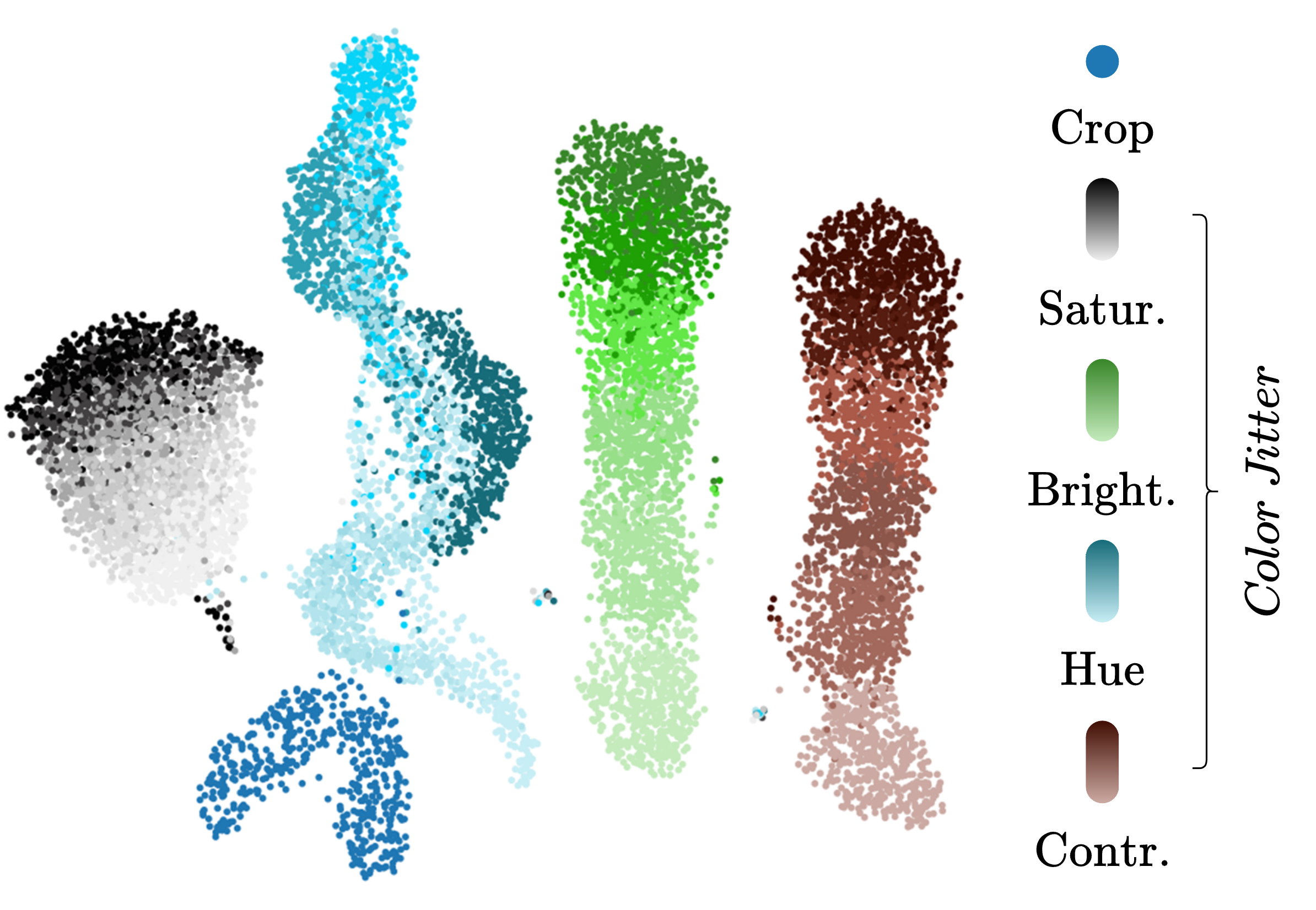}
    \caption{\small{\textbf{Visualization of Transformation Representations by Intensity.} UMAP visualization of transformation representations organized by intensity levels for each transformation type, including random crop and color jitter variations in brightness, contrast, saturation, and hue. Parameter ranges for each transformation are divided into four segments to apply varying intensities, with darker colors representing higher intensities. Representations are captured by a ResNet-18 model pretrained on STL10 with a transformation backbone.}}
    \vspace{-1em}
    \label{fig:intra_relationship}
\end{wrapfigure} 
\textbf{Transformation Representation.} To evaluate learned transformation representations of STL, we assess both parameter prediction and type classification on test images with transformations used during pretraining, specifically crop and color jitter. Parameter prediction uses MSE loss to measure accuracy in predicting transformation specifics, while type classification assesses the model’s ability to distinguish among transformation categories. As shown in Table~\ref{tab:transformation_prediction}, STL achieves the lowest MSE for crop, color jitter, and the combined metric, indicating precise capture of transformation details. In classification, STL attains 93.67\% accuracy, surpassing other models, even though AugSelf incorporates parameter regression into its learning. This demonstrates STL’s robust generalization in transformation learning without direct parameter supervision.

Additionally, a qualitative UMAP visualization of transformation representations from test image pairs reveals STL’s understanding of transformation relationships. Figure~\ref{fig:transformation_representation}(b) shows distinct clusters for each transformation type, including crop, brightness, contrast, saturation, and hue, with color-related transformations grouped closely, reflecting STL’s capture of inter-relationships among similar transformations. Figure~\ref{fig:intra_relationship} shows that transformations with similar intensity levels are positioned closer in the representation space, forming continuous representations that capture the intrinsic order of intensity within each transformation type. This structure indicates STL’s effective learning of transformation intra-relationships, demonstrating its nuanced understanding of both type and intensity.

\textbf{Transformation Equivariance.} We evaluate the accuracy of STL's learned equivariant transformations in reflecting real transformations in image space. This evaluation involves applying 60 transformations per image from the STL10 test dataset, including standalone transformations like crop and color jitter, as well as the standard combinations used during training. For each transformed representation, we rank other representations by similarity, ordering them from closest to furthest. Metrics include Mean Reciprocal Rank (MRR), Hit@k (H@k), and Precision (PRE). MRR is defined as the mean of reciprocal ranks \(1/r\), where \(r\) is the rank of the nearest representation corresponding to the designated transformation. Hit@k calculates the probability \(P(r \leq k)\) that the correct transformation rank \(r\) is within the top \(k\). PRE is the mean squared error between the representation from the top-ranked transformation and the actual transformation’s parameter vector.

As shown in Table~\ref{tab:transformation_equivariance}, STL outperforms previous methods in most metrics, indicating closer alignment of its equivariant transformations with real image transformations in representation space. Notable exceptions are crop H@5 and PRE, where other methods perform slightly better. Overall, STL captures both individual and combined transformations more faithfully, achieving robust alignment between representation and image transformations. Self-supervised transformation learning, by maintaining image-invariance, enhances input consistency for equivariant transformation learning and significantly improves alignment accuracy. Without this component, STL’s ability to capture transformation nuances declines, underscoring the role of each component in the STL framework.

\begin{table}[t]
    \renewcommand{\arraystretch}{1.1}
    \centering
    \caption{\small\textbf{Transformation Equivariance.} Evaluation of the equivariant transformation. Mean Reciprocal Rank (MRR), Hit@k (H@k), and Precision (PRE) metrics on various transformations (crop and color jitter).}
    \label{tab:transformation_equivariance}
    \resizebox{\linewidth}{!}{
        \begin{tabular}{lcccccccccccccc}
        \toprule
        & \multicolumn{4}{c}{\textbf{Crop}} & \multicolumn{4}{c}{\textbf{Color}} & \multicolumn{4}{c}{\textbf{All}}\\
        
        \cmidrule(lr){2-5} \cmidrule(lr){6-9} \cmidrule(lr){10-13}
        \textbf{Method} & MRR($\uparrow$) & H@1($\uparrow$) & H@5($\uparrow$) & PRE($\downarrow$) & MRR($\uparrow$) & H@1($\uparrow$) & H@5($\uparrow$) & PRE($\downarrow$) & MRR($\uparrow$) & H@1($\uparrow$) & H@5($\uparrow$) & PRE($\downarrow$)\\
        
        \midrule
        SEN & 0.34&0.15&0.58&0.14&0.18&0.05&0.31&3.69&0.22&0.08&0.37&2.70\\
        EquiMod & \textbf{0.37}&0.17&\textbf{0.60}&\textbf{0.13}&0.16&0.05&0.28&3.72&0.22&0.09&0.36&2.72\\
        SIE & 0.33&0.14&0.55&0.33&0.17&0.05&0.28&3.70&0.21&0.08&0.35&2.74\\
        
        \midrule
        \textbf{w/o $\mathcal{L}_\text{trans}$ (Ours)} &0.31&0.18&0.46&0.69&0.27&0.13&0.40&3.37&0.29&0.16&0.43&2.50\\
        \textbf{STL (Ours)} &\textbf{0.37}&\textbf{0.22}&0.54&0.64&\textbf{0.33}&\textbf{0.18}&\textbf{0.52}&\textbf{2.76}&\textbf{0.36}&\textbf{0.21}&\textbf{0.53}&\textbf{2.07}\\
          
        \bottomrule
        \end{tabular}
    }
\end{table}

\subsection{Ablation Studies}

\begin{table}[t]
    \tiny
    \centering
    \caption{\small{\textbf{Loss Function Ablation Study.} Image classification and transformation prediction results of ResNet-18 pretrained on STL10 with selective inclusion of loss terms for invariant learning ($\mathcal{L}_\text{inv}$), equivariant learning ($\mathcal{L}_\text{equi})$, and self-supervised transformation learning ($\mathcal{L}_\text{trans}$). For image classification, in-domain accuracy (\%) and the average accuracy (\%) across multiple out-domain datasets are shown. For transformation prediction, MSE is used for regression of crop and color transformations, and accuracy (\%) is used for transformation type classification.}}
    \label{tab:loss_function_ablation}
    \resizebox{\linewidth}{!}{
        \begin{tabular}{lccccccc}
        \toprule    
        & \multicolumn{3}{c}{\textbf{Loss Functions}} & \multicolumn{2}{c}{\textbf{Image Classification}} & \multicolumn{2}{c}{\textbf{Transformation Prediction}} \\
        \cmidrule(lr){2-4} \cmidrule(lr){5-6} \cmidrule(lr){7-8} 
        \textbf{Method} & $\mathcal{L}_\text{inv}$ & $\mathcal{L}_\text{equi}$ & $\mathcal{L}_\text{trans}$ & In-domain ($\uparrow$) & Out-domain ($\uparrow$) & Regression ($\downarrow$) & Classification ($\uparrow$) \\

        \midrule
        Only Invariance & \checkmark & - & - & 84.74 & 43.11 & 0.08 & 68.54 \\
        Only Equivariance & - & \checkmark & - & 83.53 & \textbf{49.99} & \textbf{0.02} & 93.54 \\

        \midrule
        STL w/o $\mathcal{L}_\text{inv}$ & - & \checkmark & \checkmark & 81.86 & 48.62 & \textbf{0.02} & 93.54 \\
        STL w/o $\mathcal{L}_\text{equi}$ & \checkmark & - & \checkmark & 80.99 & 47.30 & \textbf{0.02} & \textbf{93.92} \\
        STL w/o $\mathcal{L}_\text{trans}$ & \checkmark & \checkmark & - & \textbf{85.11} & 48.49 & 0.08 & 69.57 \\

        \midrule
        STL & \checkmark & \checkmark & \checkmark & 84.83 & \textbf{49.97} & \textbf{0.02} & 93.67 \\
        
        \bottomrule
        \end{tabular}
    }
\end{table}

\textbf{Loss Functions.} We conduct an ablation study to analyze the impact of each loss function on STL’s performance across image classification and transformation prediction tasks. Specifically, we examine the contributions of invariant learning (\(\mathcal{L}_\text{inv}\)), equivariant learning (\(\mathcal{L}_\text{equi}\)), and self-supervised transformation learning (\(\mathcal{L}_\text{trans}\)) by selectively removing each loss term. Table~\ref{tab:loss_function_ablation} illustrates a clear trade-off between invariance and equivariance in STL's performance. The Only Invariance configuration achieves high in-domain accuracy of 84.74\% but suffers from low out-domain accuracy of 43.11\% and limited transformation prediction capabilities, highlighting restricted generalizability. In contrast, Only Equivariance improves out-domain accuracy to 49.99\% and achieves strong transformation prediction with an MSE of 0.02, indicating enhanced generalization and transformation awareness, albeit with a slight reduction in in-domain performance. 

When \(\mathcal{L}_\text{trans}\) is omitted, the model maintains high in-domain accuracy but exhibits weak out-domain and transformation performance, suggesting that the absence of transformation representation learning leads to a focus on invariance. Excluding \(\mathcal{L}_\text{inv}\) improves out-domain accuracy to 48.62\% and enhances transformation alignment by preventing collapse into pure invariance, although this comes at a moderate cost to in-domain accuracy. Without \(\mathcal{L}_\text{equi}\), the in-domain performance decreases further as transformation learning alone lacks sufficient structure for alignment. Finally, the full STL, which incorporates all three losses, achieves the best balance, with superior performance across in-domain and out-domain tasks and optimal transformation prediction results. This configuration minimizes in-domain trade-offs while capturing a comprehensive view of transformations, ensuring alignment between representation and image transformations across both in-domain and out-domain tasks.

\begin{table}[t]
    \renewcommand{\arraystretch}{1.1}
    \centering
    \caption{\small{\textbf{Transformation Ablation Study.} Linear evaluation accuracy (\%) of ResNet-18 pretrained on STL10 with various transformations used as equivariance targets.}}
    \label{tab:transformation_ablation}
    \resizebox{\linewidth}{!}{%
        \begin{tabular}{>{\centering\arraybackslash}p{1.0cm}lccccccccccc|c}
        \toprule
        \textbf{Trans.} & \textbf{Method} & \textbf{CIFAR10} & \textbf{CIFAR100} & \textbf{Food} & \textbf{MIT67} & \textbf{Pets} & \textbf{Flowers} & \textbf{Caltech101} & \textbf{Cars} & \textbf{Aircraft} & \textbf{DTD} & \textbf{SUN397} & \textbf{Mean} \\
        
        \midrule
        \multirow{3}{*}{crop} & AugSelf & 82.89 & 54.92 & 33.19 & \textbf{39.70} & 44.40 & 64.96 & 67.63 & 15.58 & 25.38 & \textbf{41.86} & 27.89 & 45.31 \\
         & EquiMod & 83.76 & 55.33 & 32.01 & 37.76 & 41.65 & 63.00 & 66.28 & 14.18 & 24.96 & 41.54 & 26.46 & 44.27 \\
         & STL & \textbf{84.94} & \textbf{59.12} & \textbf{35.15} & 39.40 & \textbf{45.35} & \textbf{68.38} & \textbf{70.78} & \textbf{17.96} & \textbf{33.00} & \textbf{41.86} & \textbf{28.71} & \textbf{47.70} \\
        
        \midrule
        \multirow{3}{*}{color} & AugSelf & \textbf{84.33} & 57.47 & 36.57 & 39.40 & \textbf{46.80} & 71.18 & 67.91 & 17.03 & \textbf{27.12} & 43.83 & 29.37 & 47.36 \\
         & EquiMod & 82.22 & 51.77 & 31.21 & 34.18 & 39.57 & 61.17 & 62.07 & 12.51 & 21.36 & 39.52 & 23.48 & 41.73 \\
         & STL & 84.16 & \textbf{58.71} & \textbf{38.49} & \textbf{41.34} & 45.90 &\textbf{ 74.36} & \textbf{68.48} & \textbf{17.31} & \textbf{27.12} & \textbf{46.54} & \textbf{31.17} &\textbf{ 48.51} \\
        
        \midrule
        \multirow{3}{*}{\makecell{crop \\+\\color}} & AugSelf & 84.26 & 57.78 & 36.82 & 40.30 & 45.46 & 73.38 & 68.11 & 17.22 & 27.63 & 45.96 & 30.38 & 47.94 \\
         & EquiMod & 81.35 & 51.86 & 33.91 & 37.76 & 41.92 & 66.18 & 67.38 & 15.22 & 25.80 & 42.50 & 26.70 & 44.60 \\
         & STL & \textbf{85.37} & \textbf{61.05} & \textbf{39.41} & \textbf{41.27} & \textbf{46.58 }& \textbf{76.43} & \textbf{71.47} & \textbf{19.04} & \textbf{30.75} &\textbf{ 46.17} & \textbf{32.13} & \textbf{49.97} \\
        
        \midrule
        \multirow{3}{*}{all} & AugSelf & 81.76 & 54.90 & 36.51 & 40.90 & 46.17 & 71.43 & 70.14 & \textbf{18.63} & \textbf{30.96} & \textbf{45.21} & 30.40 & 47.91 \\
         & EquiMod & 84.42 & 56.65 & 34.23 & 37.99 & 42.98 & 67.16 & 68.41 & 15.18 & 26.91 & 43.94 & 26.97 & 45.89 \\
         & STL & \textbf{84.96} & \textbf{58.91} & \textbf{36.71} & \textbf{42.09} & \textbf{46.25} & \textbf{72.41} & \textbf{71.01} & 17.72 & 28.44 & 43.83 & \textbf{30.99 }& \textbf{48.48} \\
        \bottomrule
        \end{tabular}
    }
\end{table}

\textbf{Transformations.} Table~\ref{tab:transformation_ablation} shows STL’s flexibility and effectiveness across various transformations. STL achieves the highest mean accuracy across all settings, outperforming AugSelf and EquiMod in single transformations such as crop and color, as well as in combined transformations including all transformations, reaching mean accuracy scores of 47.70\%, 48.51\%, and 48.48\%, respectively. These results highlight STL’s robust equivariant learning across diverse transformation types, enabling strong generalization without constraints on specific transformations.

\textbf{Base Invariant Learning Models.} Table~\ref{tab:base_model_ablation} demonstrates STL’s compatibility with various base models such as SimCLR, BYOL, SimSiam, and Barlow Twins. STL consistently improves the mean accuracy across all base models, outperforming both AugSelf and EquiMod, with the highest overall mean accuracy achieved with BYOL at 54.44\%. These results demonstrate that STL is compatible with different invariant learning frameworks, confirming its adaptability and effectiveness regardless of the underlying representation learning method.

\begin{table}[t]
    \renewcommand{\arraystretch}{1.1}
    \centering
    \caption{\small{\textbf{Base Invariant Learning Model Ablation Study.} Linear evaluation accuracy (\%) of ResNet-18 pretrained on STL10 with various base models for invariant learning.}}
    \label{tab:base_model_ablation}
    \resizebox{\linewidth}{!}{%
        \begin{tabular}{>{\centering\arraybackslash}p{1.0cm}lccccccccccc|c}
        \toprule
        \textbf{Base} & \textbf{Method} & \textbf{CIFAR10} & \textbf{CIFAR100} & \textbf{Food} & \textbf{MIT67} & \textbf{Pets} & \textbf{Flowers} & \textbf{Caltech101} & \textbf{Cars} & \textbf{Aircraft} & \textbf{DTD} & \textbf{SUN397} & \textbf{Mean} \\
        
        \midrule
        \multirow{4}{*}{BYOL} & - & 85.55 & 59.80 & 37.54 & 42.61 & 50.61 & 73.50 & 72.46 & 23.02 & 31.71 & 44.95 & 31.63 & 50.31 \\
         & AugSelf & 87.01 & 64.84 & \textbf{43.14} & \textbf{47.24} & \textbf{52.49} & 78.88 & 75.42 & 25.47 & 37.02 & \textbf{48.03} & \textbf{34.94} & 54.04 \\
         & EquiMod & 84.64 & 56.55 & 32.74 & 39.18 & 44.64 & 66.54 & 68.37 & 15.47 & 24.27 & 42.71 & 26.96 & 45.64 \\
         & STL & \textbf{86.88} & \textbf{65.63} & 42.98 & 46.42 & 52.33 & \textbf{79.61} & \textbf{76.04} & \textbf{28.68} & \textbf{39.21} & 46.44 & 34.57 & \textbf{54.44} \\
        
        \midrule
        \multirow{4}{*}{SimSiam} & - & 83.26 & 55.69 & 34.32 & 40.52 & 46.52 & 66.06 & 69.13 & 17.15 & 27.99 & 41.91 & 28.97 & 46.50 \\
         & AugSelf & \textbf{85.44} & 62.20 & 39.78 & 43.43 & 46.77 & \textbf{77.90} & \textbf{71.72} & 18.67 & \textbf{33.30} & 45.53 & \textbf{32.65} & 50.67 \\
         & EquiMod & 81.20 & 51.23 & 31.21 & 37.99 & 40.53 & 63.98 & 64.19 & 12.22 & 22.11 & 40.69 & 25.76 & 42.83 \\
         & STL & 85.20 & \textbf{62.58} & \textbf{40.15} & \textbf{44.03} & \textbf{48.65} & 76.68 & 71.37 & \textbf{22.42} & 32.37 & \textbf{45.59} & 32.19 & \textbf{51.02} \\

        \midrule
        \multirow{4}{*}{\makecell{Barlow \\ Twins}} & - & 81.67 & 51.68 & 27.79 & 33.13 & 39.60 & 57.63 & 62.17 & 11.53 & 19.47 & 37.13 & 23.43 & 40.48 \\
         & AugSelf & 82.46 & 51.71 & 27.83 & 35.75 & 39.33 & 58.24 & 61.87 & 11.88 & 19.77 & 37.29 & 23.31 & 40.86 \\
         & EquiMod & 81.57 & 52.15 & 30.00 & 36.79 & 38.70 & 62.64 & 63.22 & 11.80 & 20.55 & 40.21 & 24.92 & 42.05 \\
         & STL & \textbf{83.74} & \textbf{56.73} & \textbf{32.69} & \textbf{38.36} & \textbf{42.65} & \textbf{67.28} & \textbf{68.09} & \textbf{16.24} & \textbf{24.33} & \textbf{41.97} & \textbf{28.53} & \textbf{45.51} \\
        \bottomrule
        \end{tabular}
    }
\end{table}

\section{Related Works}
\label{related_work}

Transformation equivariant learning captures transformation-sensitive features by embedding transformation directly into the learning process, categorized into implicit and explicit approaches. Implicit learning predicts transformations by observing changes in representations, allowing models to infer transformation without directly modeling the functions. In contrast, explicit equivariant learning encodes transformations within the representation space, enforcing behaviors in learned representations that mirror input transformations. These approaches can be reviewed in detail in the Appendix~\ref{section:supp:implicit_and_explicit}.

\textbf{Implicit Equivariant Learning.} In implicit equivariant learning, models learn to recognize applied transformations by observing changes in representations, enabling the representations to capture transformation-sensitive information. Notable methods include InfoMin~\citep{goodview}, selecting optimal views to maintain relevant task information, and Prelax~\citep{residualrelaxation}, aligning residual vectors for robust multi-view alignment. Similarly, E-SSL~\citep{e-ssl} and AugSelf~\citep{augself} leverage transformation-aware auxiliary tasks to train models to preserve transformation-sensitive details, enhancing robustness by maintaining sensitivity to transformation variance.

\textbf{Explicit Equivariant Learning.}
On the other hand, explicit equivariant learning directly encodes input transformation into the representation space, building equivariant transformations that operate consistently in representation space. AEAE~\citep{affine} leverages group actions to embed transformation effects explicitly in the representation space, while SymReg~\citep{groupinvariants} enhances transformation consistency by selecting optimal loss terms based on transformation group information. CARE~\citep{structuring} introduces rotational symmetry by aligning embeddings directly with input rotations, providing robust transformation encoding. Similarly, SEN~\citep{sen} applies symmetric embedding networks to synchronize transformations in the input space with learned representations. In a more flexible approach, EquiMod~\citep{equimod} models equivariant transformations by conditioning transformation labels as inputs and dynamically adapting representations through a neural network. Building on these approaches, SIE~\citep{sie} separates invariant and equivariant representations, using dedicated networks to distinctly capture transformation-sensitive and invariant aspects.

\textbf{Applications of Equivariant Learning.} The applicability of equivariant learning extends across various domains, including robotics, medical imaging, molecular modeling, and multimodal representation learning. By leveraging inherent symmetries within data, it enhances sample efficiency in robotic manipulation~\citep{wang2022robot}, improves accuracy in medical image processing~\citep{hashemi2023spherical}, and boosts predictive performance in molecular data analysis~\citep{winter2022unsupervised}. In multimodal contexts, it achieves finer alignment by considering transformations within each modality, facilitating robust cross-modal understanding~\citep{uniclip, uniav}. These applications underscore the broad utility of equivariant learning in aligning model representations with domain-specific transformations.

\section{Discussion and Conclusion}
\label{discussion_and_conclusion}

The Self-supervised Transformation Learning (STL) approach introduces a novel method for deriving transformation representations that capture equivariance without relying on predefined transformation labels. This framework optimizes representational versatility through a synergy of three key loss functions: invariant learning, equivariant learning, and self-supervised transformation learning. These loss functions allow representations to adapt to complex transformation dynamics, including interdependencies among transformations, thereby greatly enhancing model generalization. STL consistently outperforms existing methods across 7 out of 11 classification tasks and demonstrates exceptional performance in object detection, proving its strong ability to generalize across diverse transformations. Additionally, integrating STL with AugMix yields robust performance improvements across all tasks, demonstrating enhanced resilience. STL’s adaptability and consistent performance across various transformations and base models underscore its versatility for broad applications.

\textbf{Limitations.} 
While STL significantly advances equivariant learning, it encounters challenges with transformations that extend beyond single image pairs. Complex transformations, such as those involving combinations or mixtures of multiple images (e.g., mixup~\citep{mixup}), fall outside STL's current capacity as it relies on pairwise transformation representations. Further research could explore ways to adapt STL to accommodate more complex, multi-image transformations and better capture their inherent structure.

\textbf{Broader Impacts.} 
STL holds promise for applications requiring precise and interpretable transformations, such as in medical imaging analysis and autonomous driving. However, as STL learns transformation representations from the data, it may inherit biases embedded in the training data, raising fairness concerns, especially in sensitive domains. Implementing fairness-aware training techniques and thorough validation processes could help mitigate these risks. Additionally, while STL advances model robustness and generalization, its computational demands may have environmental implications. Efficiency improvements, such as model distillation, could reduce the model’s energy footprint, supporting sustainable deployment.

\section*{Acknowledgments}
This work was supported by Institute of Information \& Communications Technology Planning \& Evaluation(IITP) grant funded by the Korea government(MSIT) (RS-2024-00439020, Developing Sustainable, Real-Time Generative AI for Multimodal Interaction, SW Starlab).

{\small
\bibliographystyle{abbrvnat}
\bibliography{egbib}
}

\newpage
\appendix

\label{appendix}

\section{STL Formulations for Various Base Invariant Models}
\label{section:supp:various_base_models}

\subsection{STL Extension on BYOL}
In adapting STL to BYOL~\citep{byol}, we utilize dissimilarity loss of BYOL to define the invariant, equivariant, and transformation losses. BYOL’s dissimilarity metric is:
\begin{equation}
    \mathcal{L}_\text{BYOL}(y, y^+; g, q, \theta, \xi) = \|\overline{q}_\theta (g_\theta (y)) - \overline{g}_\xi (y^+) \|_2^2,
\end{equation}
where \(g_\theta\) denotes the projection network parameterized by \(\theta\), \(q_\theta\) is the prediction network also parameterized by \(\theta\), and \(g_\xi\) represents the target network parameterized by \(\xi\). The terms \(\overline{q}_\theta\) and \(\overline{g}_\xi\) refer to the normalized outputs of \(q_\theta\) and \(g_\xi\), respectively. 
Using this, we define the STL objectives as follows:
\begin{align}
    &\mathcal{L}_\text{inv}(x, t) =  \mathcal{L}_\text{BYOL}(f(x), f(t(x)); g_\text{inv}, q_\text{inv}, \theta, \xi),\\
    &\mathcal{L}_\text{equi}(x, x', t) =  \mathcal{L}_\text{BYOL}(\phi(y^{x'}_t, f(x)), f(t(x)); g_\text{equi}, q_\text{equi}, \theta, \xi),\\
    &\mathcal{L}_\text{trans}(x, x', t) =  \mathcal{L}_\text{BYOL}(y^x_t, y^{x'}_t; g_\text{trans}, q_\text{trans}, \theta, \xi).
\end{align}

\subsection{STL Extension on SimSiam}
For SimSiam~\citep{simsiam}, STL uses SimSiam’s dissimilarity loss:
\begin{equation}
    \mathcal{L}_\text{SimSiam}(y, y^+; g, h) = \frac{1}{2} \mathcal{D}(h(g(y)), \text{stopgrad}(g(y^+))) + \frac{1}{2} \mathcal{D}(h(g(y^+)), \text{stopgrad}(g(y))),
\end{equation}
where \(g\) denotes the projection network, \(h\) is the prediction network, \(\text{stopgrad}\) indicates an operation that halts gradient backpropagation, and \(\mathcal{D}\) represents cosine similarity.
This enables us to structure the STL losses as:
\begin{align}
    &\mathcal{L}_\text{inv}(x, t) =  \mathcal{L}_\text{SimSiam}(f(x), f(t(x)); g_\text{inv}, h_\text{inv}),\\
    &\mathcal{L}_\text{equi}(x, x', t) =  \mathcal{L}_\text{SimSiam}(\phi(y^{x'}_t, f(x)), f(t(x)); g_\text{equi}, h_\text{equi}),\\
    &\mathcal{L}_\text{trans}(x, x', t) =  \mathcal{L}_\text{SimSiam}(y^x_t, y^{x'}_t; g_\text{trans}, h_\text{trans}).
\end{align}

\subsection{STL Extension on Barlow Twins}
For Barlow Twins~\citep{barlow}, STL applies Barlow Twins’ dissimilarity loss:
\begin{equation}
    \mathcal{L}_\text{BarlowTwins}(Y=\{y_i\}_i, Y^+=\{y^+_i\}_i; g, \lambda) = \sum_i (1 - \mathcal{C}_{ii})^2 + \lambda \sum_{i} \sum_{j \neq i} \mathcal{C}_{ij}^2,
\end{equation}
where \(\mathcal{C}\) denotes the cross-correlation matrix between embeddings of transformed views \(Y\) and \(Y^+\), and \(\lambda\) is a regularization parameter that controls the weight of off-diagonal terms in \(\mathcal{C}\), penalizing redundancy in the representations. Define \(X = \{x_i\}_i\), \(X' = \{x'_i\}_i\), and \(T = \{t_i\}_i\), representing the sets of input images, paired images, and transformations, respectively. Then, the STL losses for Barlow Twins are:
\begin{align}
    &\mathcal{L}_\text{inv}(X, T) =  \mathcal{L}_\text{BarlowTwins}(\{f(x_i)\}_i, \{f(t_i(x_i))\}_i; g_\text{inv}, \lambda_\text{inv}),\\
    &\mathcal{L}_\text{equi}(X, X', T) =  \mathcal{L}_\text{BarlowTwins}(\{\phi(y^{x'_i}_{t_i}, f(x_i))\}_i, \{f(t_i(x_i))\}_i; g_\text{equi}, \lambda_\text{equi}),\\
    &\mathcal{L}_\text{trans}(X, X', T) =  \mathcal{L}_\text{BarlowTwins}(\{y^{x_i}_{t_i}\}_i, \{y^{x'_i}_{t_i}\}_i; g_\text{trans}, \lambda_\text{trans}).
\end{align} 

\newpage
\section{Datasets}
\label{section:supp:dataset_info}
\begin{table}[h]
\centering
\caption{\small\textbf{Dataset Information.} Overview of dataset composition and evaluation metrics. Each dataset specifies the number of classes, training/validation/test splits, and the corresponding evaluation metric.}
\label{table:dataset_info}
\resizebox{\textwidth}{!}{
    \begin{tabular}{llrrrrc}
        \toprule
        \textbf{Category} & \textbf{Dataset} & \textbf{\# of classes} & \textbf{Training} & \textbf{Validation} & \textbf{Test} & \textbf{Metric} \\
        
        \midrule
        \multirow{2}{*}{(a) Pretraining}
        & STL10 \citep{stl10}               & 10  & 105,000 & - & - & -  \\
        & ImageNet100 \citep{imagenet, imagenet_samplesplit} & 1,000 & 126,689 & - & - & - \\
        
        \midrule
        \multirow{12}{*}{(b) Linear Evaluation}
        & CIFAR10 \citep{cifar}         &  10 & 45,000 & 5,000 & 10,000 & Top-1 accuracy \\
        & CIFAR100 \citep{cifar}        & 100 & 45,000 & 5,000 & 10,000 & Top-1 accuracy \\
        & Food    \citep{food}       & 101 & 68,175 & 7,575 & 25,250 & Top-1 accuracy \\
        & MIT67 \citep{mit67}             &  67 &  4,690 &  670 &  1,340 & Top-1 accuracy \\
        & Pets \citep{pets}                 &  37 &  2,940 &  740 &  3,669 & Mean per-class accuracy \\
        & Flowers \citep{flowers} & 102 &  1,020 & 1,020 &  6,149 & Mean per-class accuracy \\
        & Caltech101 \citep{caltech101}        & 101 &  2,525 &  505 &  5,647 & Mean Per-class accuracy \\
        & Cars \citep{cars}            & 196 &  6,494 & 1,650 &  8,041 & Top-1 accuracy \\
        & Aircraft \citep{aircraft}      & 100 &  3,334 & 3,333 &  3,333 & Mean Per-class accuracy \\
        & DTD (split 1) \citep{dtd}         & 47  &  1,880 & 1,880 &  1,880 & Top-1 accuracy \\
        & SUN397 (split 1) \citep{sun397}        & 397 & 15,880 & 3,970 & 19,850 & Top-1 accuracy \\
        
        \midrule
        {(c) Object Detection}
        & VOC2007+2012 \citep{voc07} & 20 & 16,551 & - & 4,952 & Average Precision \\
        \bottomrule
    \end{tabular}}
    \vspace{1em}
\end{table}

Table \ref{table:dataset_info} presents detailed descriptions of (a) pre-taining dataset, (b) linear evaluation datasets, and (c) object detection and instance segmentation dataset.
For linear evaluation dataset, validation samples are randomly selected from the training split if an official validation split is not provided.
For the object detection and instance segmentation dataset, we use \texttt{trainval} set for training VOC07+12~\citep{voc07} while only using \texttt{test} set for testing.

\section{Evaluation Protocols}
\label{section:supp:evaluation_protocol}

\textbf{Linear Evaluation.} Following established protocols~\citep{eval_protocol, simclr, byol}, we train linear classifiers on frozen representations from center-cropped images resized to $224\times224$ (or $96\times96$ for STL10). No data augmentation is applied. Each image is resized along its shorter side to 224 and then center-cropped to $224\times224$. The $\ell_2$-regularized cross-entropy objective is minimized using L-BFGS, with regularization selected from 45 logarithmically spaced values between $10^{-6}$ and $10^5$ on the validation set. The optimal model is then retrained on both training and validation splits, with test accuracy reported. L-BFGS is capped at 5,000 iterations, with each step initialized from the previous solution.

\textbf{Object Detection.} We use the VOC2007+2012 \texttt{trainval} set with 16,551 images to train a Faster R-CNN~\citep{fastrcnn} with a C-4 backbone. Training spans 24,000 iterations with a batch size of 16 and SyncBatchNorm. The learning rate starts at 0.1, decreasing by a factor of 10 at 18,000 and 22,000 iterations. A linear warmup~\citep{appendix_1} is applied for the first 1,000 iterations with a 0.333 slope.

\section{Transformation Labels}
\label{section:supp:transformation_labels}
In this section, we describe the labels of the transformations utilized in AugSelf~\citep{augself} (random crop, horizontal flip, color jitter, grayscale, and Gaussian blur) in the following.
The labels are designed by the parameters in each transformation.
\begin{itemize}
    \setlength\itemsep{0.02em}
    \item \textbf{\texttt{RandomResizedCrop}}. The labels are constructed with the center points, $H_{center}$ and $W_{center}$, of the crop and the height $H$ and width $W$ values of the cropping area.
    \item \textbf{\texttt{RandomHorizontalFlip}}. As flip transformation is a simple operation, the label is 0 or 1.
    \item \textbf{\texttt{ColorJitter}}. The four parameters of color jitter are brightness, contrast, saturation, and hue. Each parameter has its own range, with brightness, contrast, and saturation ranging from 0.0 to 1.0, while hue ranges from 0.0 to 0.5.
    \item \textbf{\texttt{RandomGrayscale}}. This transformation applies the grayscale to the image. In line with the flip, the label is 0 or 1.
    \item \textbf{\texttt{GaussianBlur}}. The standard deviation is utilized for the Gaussian blur label ranging from 0.1 to 2.0.
\end{itemize}

\section{Pretraining Setups}
\label{section:supp:pretraining_setup}
For the pretraining experiments, we use NVIDIA RTX4090.

\subsection{ImageNet100 Pretraining}\label{section:supp:pretrain_imagenet100}

We conduct pretraining on the ResNet-50 architecture~\citep{resnet18} using ImageNet100, a subset of ImageNet containing 100 categories~\citep{imagenet}, with dataset splits consistent with those in~\citep{imagenet_samplesplit}. All methods are trained for 500 epochs with a batch size of 256, using a cosine learning rate schedule without restarts~\citep{sgdr}. The initial learning rate is set at 0.03, with a weight decay of 0.0005. The model includes a 3-layer projection MLP head, \( g(\cdot) \), with a hidden dimension of 2048 and an output dimension of 128. Batch normalization~\citep{bn} is excluded from the last layer.

\subsection{STL10 Pretraining}\label{section:supp:pretrain_stl10}

For pretraining on the STL10 dataset~\citep{stl10}, we use the standard ResNet-18 architecture~\citep{resnet18}. All methods utilize stochastic gradient descent (SGD) with a learning rate of 0.03, a batch size of 256, a weight decay of 0.0005, and a momentum of 0.9. The learning rate follows a cosine decay schedule without restarts~\citep{sgdr}.

\textbf{SimCLR}~\citep{simclr}. A 3-layer projection MLP head \( g(\cdot) \) with a hidden dimension of 512 and an output dimension of 128 is used, with batch normalization excluded from the final layer. In contrastive learning, we apply a temperature scaling parameter of 0.2.

\textbf{Barlow Twins}~\citep{barlow}. A 2-layer projection MLP head \( g(\cdot) \) is employed, with a hidden dimension of 512 and an output dimension of 2048. Batch normalization is excluded from the last layer.

\textbf{BYOL}~\citep{byol}. The model uses a 2-layer projection MLP head \( g(\cdot) \), with a hidden dimension of 4096 and an output dimension of 256, omitting batch normalization in the final layer.

\textbf{SimSiam}~\citep{simsiam}. We employ a 2-layer projection MLP head \( g(\cdot) \) with both hidden and output dimensions of 2048, with batch normalization excluded from the final layer.

\section{Image Classification Results of STL10-pretrained Models}
\label{section:supp:stl10_results}
\begin{table}[h]
    \renewcommand{\arraystretch}{1.1}
    \centering
    \caption{\small{\textbf{Image Classification.} Evaluation of representation on in-domain and 11 downstream out-domain classification task: Linear evaluation accuracy (\%) of ResNet-18~\citep{resnet18} pretrained on STL10~\citep{stl10} averaged over three random seeds (mean $\pm$ std).}}
    \tiny
    \label{tab:stl10_results}
    \resizebox{\linewidth}{!}{
        \begin{tabular}{l|c|ccccccccccc|c}
        \toprule
        & \textbf{In-domain} & \multicolumn{12}{c}{\textbf{Out-domain}}\\
        \textbf{Method} & \textbf{STL10} & \textbf{CIFAR10} & \textbf{CIFAR100} & \textbf{Food} & \textbf{MIT67} & \textbf{Pets} & \textbf{Flowers} & \textbf{Caltech101} & \textbf{Cars} & \textbf{Aircraft} & \textbf{DTD} & \textbf{SUN397} & \textbf{Mean} \\
        
        \midrule
        \multicolumn{13}{l|}{\textit{\textcolor{gray}Transformation Invariant Learning} :} \\
        SimCLR & 84.74$\pm$0.18 & 80.89$\pm$2.70 & 51.12$\pm$2.50 & 32.23$\pm$0.18 & 37.61$\pm$1.13 & 44.10$\pm$0.38 & 63.55$\pm$0.67 & 66.17$\pm$0.55 & 14.44$\pm$0.21 & 24.13$\pm$0.84 & 40.32$\pm$0.33 & 26.23$\pm$0.10 & 43.72$\pm$0.65 \\
        with AugMix & \textbf{85.58$\pm$0.16} & 81.72$\pm$0.75 & 52.42$\pm$0.36 & 33.22$\pm$0.31 & 39.63$\pm$0.98 & 45.16$\pm$0.78 & 65.32$\pm$0.68 & 69.81$\pm$0.30 & 15.66$\pm$0.74 & 25.79$\pm$1.00 & 42.06$\pm$0.55 & 27.95$\pm$0.06 &	45.34$\pm$0.21 \\
        
        \midrule
        \multicolumn{13}{l|}{\textit{\textcolor{gray}Implicit Transformation Equivariant Learning} :} \\
        E-SSL & 85.19$\pm$0.08 & 82.82$\pm$2.17 & 54.89$\pm$2.98 & 35.11$\pm$0.16 & 39.55$\pm$0.20 & 45.06$\pm$0.15 & 69.19$\pm$1.47 & 68.23$\pm$0.65 & 16.51$\pm$0.23 & 26.84$\pm$0.80 & 43.87$\pm$0.82 & 28.80$\pm$0.06 & 46.44$\pm$0.33 \\
        AugSelf & 84.99$\pm$1.05 & 84.12$\pm$0.94 & 57.59$\pm$0.80 & 36.63$\pm$0.08 & 40.90$\pm$0.97 & 46.09$\pm$0.44 & 72.45$\pm$1.53 & 69.58$\pm$0.27 & 17.58$\pm$0.17 & 27.73$\pm$0.43 & 44.24$\pm$1.68 & 30.49$\pm$0.35 & 47.94$\pm$0.25 \\
        
        \midrule
        \multicolumn{13}{l|}{\textit{\textcolor{gray}Explicit Transformation Equivariant Learning} :} \\
        SEN     & 78.66$\pm$0.40 & 81.01$\pm$0.57 & 51.59$\pm$1.10 & 30.00$\pm$0.23 & 34.00$\pm$1.26 & 35.76$\pm$0.28 & 60.93$\pm$1.18 & 64.35$\pm$0.16 & 12.12$\pm$0.51 & 24.87$\pm$1.32 & 38.12$\pm$0.31 & 23.24$\pm$0.30 & 41.45$\pm$0.32 \\
        EquiMod & 84.28$\pm$0.18 & 83.63$\pm$1.59 & 55.94$\pm$2.34 & 34.01$\pm$0.49 & 38.78$\pm$0.63 & 42.94$\pm$0.86 & 66.75$\pm$0.69 & 68.70$\pm$0.87 & 15.49$\pm$0.72 & 26.84$\pm$1.11 & 43.03$\pm$1.05 & 27.41$\pm$0.43 & 45.77$\pm$0.79 \\
        SIE & 83.60$\pm$0.13 & 82.60$\pm$1.89 & 53.43$\pm$2.59 & 32.97$\pm$0.27 & 37.14$\pm$0.99 & 41.03$\pm$0.25 & 65.24$\pm$1.07 & 67.07$\pm$0.29 & 14.04$\pm$0.54 & 25.91$\pm$0.68 & 42.04$\pm$0.35 & 25.97$\pm$0.12 & 44.31$\pm$0.64 \\
        
        \textbf{STL (Ours)} & 84.83$\pm$0.21 & 85.22$\pm$0.17 & 60.13$\pm$0.81 & 38.05$\pm$1.19 & 43.53$\pm$2.04 & 46.57$\pm$0.56 & 73.50$\pm$2.58 & 71.36$\pm$0.20 & 18.85$\pm$0.17 & 30.25$\pm$1.00 & 45.34$\pm$0.73 & 31.63$\pm$0.44 & 49.49$\pm$0.42 \\
        \textbf{with AugMix} & \textbf{85.57$\pm$0.19} & \textbf{86.01$\pm$0.47} & \textbf{62.07$\pm$0.25} & \textbf{40.16$\pm$0.13} & \textbf{44.90$\pm$1.00} & \textbf{46.69$\pm$0.23} & \textbf{77.37$\pm$0.79} & \textbf{73.29$\pm$0.19} & \textbf{19.32$\pm$0.48} & \textbf{30.87$\pm$1.04} & \textbf{48.71$\pm$0.67} & \textbf{33.44$\pm$0.20} & \textbf{51.17$\pm$0.12} \\
        \bottomrule
        \end{tabular}
    }
\end{table}

\newpage
\section{Explicit and Implicit Equivariant Learning}
\label{section:supp:implicit_and_explicit}
\begin{figure}[h]
    \centering
    \includegraphics[width=\textwidth]{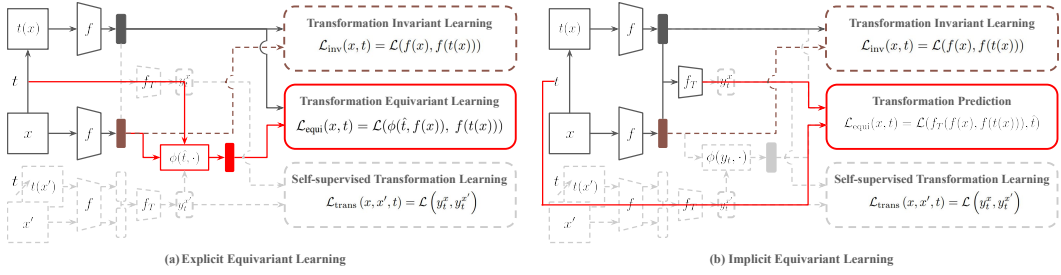}
    \caption{\small{\textbf{Explicit and Implicit Equivariant Learning.} Transformation equivariant learning with transformation labels is divided into (Left) explicit and (Right) implicit equivariant learning.}}
    \label{fig:explicit_and_implicit}
\end{figure}

Figure~\ref{fig:explicit_and_implicit} illustrates the framework of explicit and implicit equivariant learning. Explicit methods like SEN~\citep{sen}, EquiMod~\citep{equimod}, and SIE~\citep{sie} apply a direct equivariant transformation network to representations, leveraging transformation labels for alignment. In contrast, implicit methods, such as E-SSL~\citep{e-ssl} and AugSelf~\citep{augself}, infer transformation effects without directly applying transformations by utilizing auxiliary tasks to deduce transformation states.

\end{document}